\documentclass[conference]{IEEEtran}

\IEEEoverridecommandlockouts

\usepackage{cite}
\usepackage{amsmath,amssymb,amsfonts}
\usepackage{algorithmic}
\usepackage{graphicx}
\usepackage{textcomp}
\usepackage{xcolor}
\usepackage{booktabs}
\usepackage{multirow}
\usepackage{makecell}
\usepackage{url}
\usepackage{hyperref}
\usepackage{cuted}
\newcommand{\ignore}[1]{}

\def\BibTeX{{\rm B\kern-.05em{\sc i\kern-.025em b}\kern-.08em
    T\kern-.1667em\lower.7ex\hbox{E}\kern-.125emX}}
\begin{document}

\title{

Label Semantics for Robust \\Hyperspectral Image
Classification

\vspace{-4mm}

}

\author{

Rafin Hassan\textsuperscript{1}, Zarin Tasnim Roshni\textsuperscript{1}, Rafiqul Bari\textsuperscript{1}, Alimul Islam\textsuperscript{1},\\ Nabeel Mohammed\textsuperscript{1},
Moshiur Farazi\textsuperscript{2}, Shafin Rahman\textsuperscript{1} \\
\textsuperscript{1}Department of Electrical and Computer Engineering,
North South University, Dhaka, Bangladesh \\
\textsuperscript{2}University of Doha for Science and Technology, Doha, Qatar \\
\textsuperscript{1}\{rafin.hassan, zarin.roshni, rafiqul.bari, alimul.islam, nabeel.mohammed,  shafin.rahman\}@northsouth.edu \\
\textsuperscript{2}moshiur.farazi@udst.edu.qa

\vspace{-5.5mm}

}

\maketitle

\begin{abstract}
Hyperspectral imaging (HSI) classification is a critical tool with widespread applications across diverse fields such as agriculture, environmental monitoring, medicine, and materials science.
Due to the limited availability of high-quality training samples and the high dimensionality of spectral data, HSI classification models are prone to overfitting and often face challenges in balancing accuracy and computational complexity. 
Furthermore, most of HSI classification models are mono-modal, where it solely relies on spectral-spatial data to learn decision boundaries in the high dimensional embedding space. 
To address this, we propose a general-purpose Semantic Spectral-Spatial Fusion Network (S3FN) that uses contextual, class-specific textual descriptions to complement the training of an HSI classification model.
Specifically, S3FN leverages LLMs to generate comprehensive textual descriptions for each class label that captures their unique characteristics and spectral behaviors. 
These descriptions are then embedded into a vector space using a pre-trained text encoder such as BERT or RoBERTa to extract meaningful label semantics which in turn leads to a better feature-label alignment for improved classification performance. 
To demonstrate the effectiveness of our approach, we evaluate  our model on three diverse HSI benchmark datasets — Hyperspectral Wood , HyperspectralBlueberries, and DeepHS-Fruit and report significant performance boost.
Our results highlight the synergy between textual semantics and spectral-spatial data, paving the way for further advancements in semantically augmented HSI classification models. Codes are be available in: \url{https://github.com/milab-nsu/S3FN}
\end{abstract}

\begin{IEEEkeywords}
Hyperspectral Image Classification, LLM-Generated Class Descriptions, Textual Guidance.
\end{IEEEkeywords}

\section{Introduction}

Hyperspectral imaging (HSI) classification is a vital technique with applications in diverse fields such as agriculture, environmental monitoring, and material science. Unlike conventional imaging, which captures only three spectral bands (RGB), HSI acquires hundreds of contiguous spectral bands, providing rich spectral information for precise material differentiation. This detailed spectral data enables fine-grained analysis, allowing materials to be identified based on their unique spectral signatures. However, effectively leveraging high-dimensional spectral information poses significant challenges and requires advanced methods to extract meaningful features. In this paper, we introduce a novel approach to enhance HSI classification by employing automatically generated semantic descriptions of class labels.

\begin{figure}[!t]
    \raggedleft
    \includegraphics[width=\linewidth]{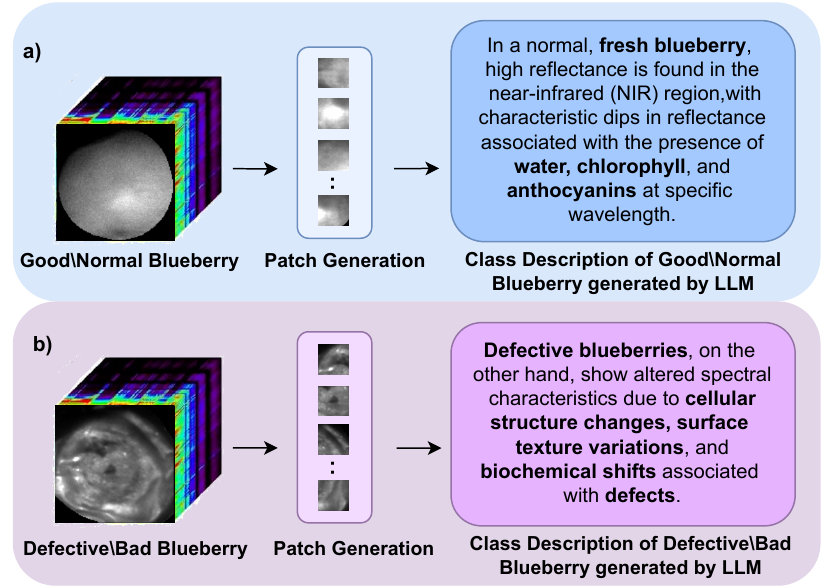}
    \caption{Quality assessment \textbf{(\textit{Good/Defective})} of blueberries using hyperspectral imaging (HSI). \textbf{(a)} A \textbf{\textit{Good}} blueberry exhibits high reflectance in the near-infrared (NIR) region, with characteristic dips corresponding to water, chlorophyll, and anthocyanins at specific wavelengths. \textbf{(b)} A
    \textbf{\textit{Defective}} Blueberry shows altered spectral patterns due to cellular structure changes, surface texture variations, and biochemical shifts. The  generated patches preserve key regions in the hyperspectral images, thus enabling the LLM generated descriptions of differences to be relevant to the patch.
    }
    \label{fig:Inro}
    
    \vspace{-6mm}
    
\end{figure}

Existing methods for HSI classification often rely solely on spectral-spatial features extracted from the data. Although techniques like 2D and 3D CNNs have shown effectiveness in capturing spatial and spectral information, they often overlook the semantic relationships between different classes \cite{simplified_2D_3DCNN_paper, Urban_Feature_Extraction_paper, HSIC_2d_3DCNN_paper, orange_paper}. Pan et al. \cite{HSIC_ErtingPan} addressed the challenge of HSI classification models that struggle to generalize to new data sets. HSI classification methods are typically trained and evaluated on a single dataset, limiting their applicability to new scenarios. They proposed a new paradigm where classifiers are trained on one dataset and tested on another, incorporating zero-shot learning to deal with unseen categories in the testing set. Their approach involves using word-semantic representations to establish associations between seen and unseen categories. However, this work relies on pre-trained word vectors, such as Word2Vec, that may not accurately capture the subtle spectral differences between HSI categories. Categories or classes with similar names might be semantically similar on the basis of their word vector representations but have distinct spectral characteristics. The use of pre-trained encoders (Word2Vec) does not capture this difference. The reliance on potentially inaccurate and static word vector representations could limit the model's ability to differentiate between semantically similar but spectrally different classes. He et al. \cite{LPFormer} proposed LPFormer, a learnable prompt-based transformer model for domain generalization in HSI classification. Their method utilizes learnable prompt words to align visual and textual features, addressing the challenges of prompt engineering in HSI classification. However, LPFormer relies on relatively simple prompts, primarily using class names, which may not provide sufficient contextual information for an accurate classification. Furthermore, their proposed model requires training both a ViT and an encoder, which is computationally very expensive, especially when dealing with high-dimensional data such as HSI. Both \cite{HSIC_ErtingPan} and \cite{LPFormer}, while incorporating semantic information, do not utilize detailed class descriptions to generate semantic embeddings. This absence of rich semantic context from class descriptions limits the models' ability to fully leverage semantic information for improved classification. This gap highlights the need for approaches that can effectively incorporate comprehensive semantic information derived from detailed class descriptions to enhance the feature representation and improve classification performance. To bridge this gap, Fig. \ref{fig:Inro} demonstrates how LLM-generated class descriptions relate to key features of good and defective blueberries.

The proposed method addresses these limitations by incorporating LLM-guided semantic label embeddings to enrich the feature representation of HSI data. This approach involves utilizing a LLM to generate comprehensive textual descriptions of each class label, capturing their unique characteristics and spectral behaviors. By using an LLM, the method avoids several potential issues associated with alternative approaches, such as inconsistent, incomplete, or ambiguous manual descriptions, which may lack the necessary detail and domain-specific knowledge. Unlike traditional static word embeddings, LLM-generated descriptions provide rich semantic information that enhances class embeddings with deeper insights into spectral relationships. These descriptions are then encoded into semantic embeddings using pre-trained text encoders like BERT or RoBERTa, which, by using their self-attention mechanism, capture the contextual relationships within the class descriptions, allowing for a more dynamic understanding of class characteristics. This rich semantic context improves the model’s ability to generalize. By integrating these context-enriched semantic embeddings with spectral-spatial features extracted by a 3D-CNN, our model achieves enhanced feature-label alignment, enabling improved classification accuracy. The use of LLM-guided descriptions helps enable a more robust and generalizable model. The contributions of this paper are as follows:

\begin{itemize}\vspace{-2pt}

    \item We used LLM-generated class descriptions as textual guidance for generating semantic label embeddings that describe the Hyperspectral Images.

    \item A fusion architecture, \textbf{Semantic Spectral-Spatial Fusion Network (S3FN)}, which aligns the HSI image with the embeddings of the semantic label/description.
    
    \item The model is evaluated in three diverse HSI datasets, Hyperspectral Images for Wood Recognition\cite{wood_dataset}, HyperspectralBlueberries \cite{blueberry_dataset}, and DeepHS-Fruit\cite{ripeness_paper_dataset}.
    
\end{itemize}

\section{Related Works}

\noindent \textbf{Different Tasks With HSI:} HSI has been used in numerous fields. Crop classification and monitoring techniques use HSI to identify and classify various crops, including wheat\ignore{\cite{ozdougan2025wheat, tyagi2025nondestructive}}, rice\ignore{\cite{zhai2025simultaneously}}, corn\ignore{\cite{aneece2024machine, huang2024detection}}, soybean\ignore{\cite{yang2024classification}}, sorghum\ignore{\cite{zhao2024rapid}}, and alfalfa\ignore{\cite{aneece2024machine}}. These studies often involve machine learning algorithms to analyze spectral data.
In quality assessment, agricultural items including peanuts \cite{yang2024rapid} corn seeds \cite{huang2024detection}, and wheat grains \cite{ozdougan2025wheat} are evaluated using HSI\ignore{\cite{yang2024rapid, zhao2024multilayer}, and forage \cite{peng2024estimation}}. This includes detecting mechanical damage, identifying imperfect grains, and determining the botanical composition of the forage.
In the detection of diseases and contaminants, research is conducted on the detection of contaminants such as Cladosporium fulvum in tomatoes \cite{zhao2024early}, as well as aflatoxin in corn \cite{tao2024raman} and peanuts \cite{zhao2024multilayer}.
In monitoring growth and physiological parameters, HSI is used to monitor plant growth, water content, and chlorophyll content in rice \cite{zhai2025simultaneously} and to determine the moisture content of edamame \cite{li2024detection}.    
In weed identification, HSI is used to identify and classify different weed species, often in a time series context, to understand the growth stages \cite{ronay2024weed}\ignore{\cite{ronay2024weed, ram2024weedcube}}.
The technology is also applied to classify coated maize seed vigor \cite{wonggasem2024utilization}, evaluate the quality of Chinese medicine \cite{zhong2024intelligent}, detect the lactic acid content of maize silage \cite{xue2024rapid}, and evaluate the reduced sugar content, moisture and hollow rate in red ginseng \cite{bai2024hyperspectral}.
Moreover, in the field of remote sensing and environmental monitoring, HSI plays a very significant role in many tasks. Hyperspectral data analysis focuses on the development of new methods and algorithms to analyze hyperspectral data, including feature extraction \cite{rs16173230}\ignore{\cite{articleUn, articleMaggu, min14090923, rs16173230}}, dimensionality reduction, and data fusion techniques \cite{zhao2024early}\ignore{\cite{zhao2024early, raj2023investigating}}. It also includes the development of tools to handle and process hyperspectral data \cite{barbato2024ticino}\ignore{\cite{zhou2025hytasx, wang2024noise, barbato2024ticino, daniel2024enhanced, zhang2024spatial, qu2024compressed}}. It was also used to collect atmospheric cloud datasets and improve pixel classification \cite{yan2024collection}. HSI research explores the combination of hyperspectral data with other data sources, such as radar, Landsat, and LiDAR, to enhance analysis \cite{raj2023investigating}.
In this paper, we focus on three specific HSI classification problems: distinguishing wood types (material classification), identifying defective blueberries (quality assessment and plant disease detection), and determining fruit ripeness (growth monitoring) from the Hyperspectral Images.

\begin{figure*}[!t]

    \centering
    
    \includegraphics[width=\textwidth]{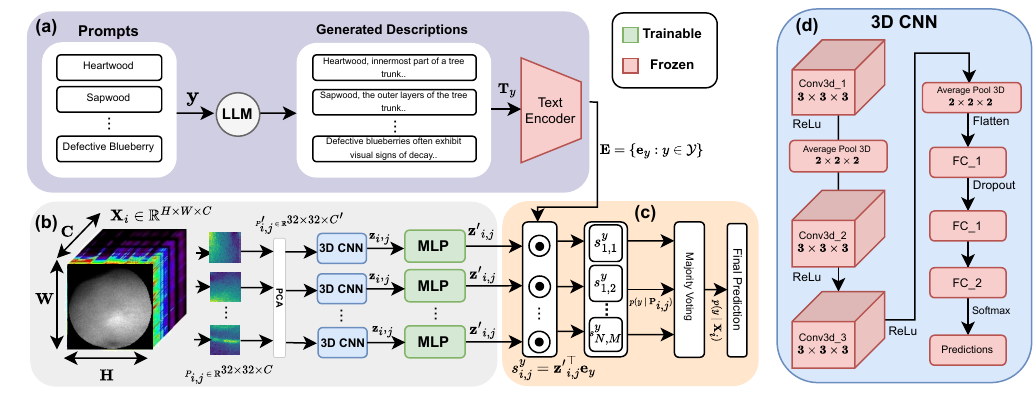}
    
     \caption{Proposed S3FN Architecture. The Semantic Spectral-Spatial Fusion Network (S3FN) integrates spectral-spatial feature extraction with semantically enriched label embeddings for robust hyperspectral image (HSI) classification through four key stages: (a) For each class $y \in \mathcal{Y}$, comprehensive textual descriptions $T_y$ generated via LLM prompts (as shown in Table~\ref{tab:description_table}) are encoded into semantic label embeddings $\mathbf{E} = \{\mathbf{e}_y \in \mathbb{R}^d \mid y \in \mathcal{Y}\}$ using pre-trained text encoders (e.g., BERT or RoBERTa). (b) Each HSI image $\mathbf{X}_i \in \mathbb{R}^{H \times W \times C}$ is partitioned into $M$ non-overlapping patches $\mathbf{P}^i_j \in \mathbb{R}^{32 \times 32 \times C}$, followed by global PCA to reduce spectral dimensionality to $C' \ll C$, yielding compressed patches \(\mathbf{P'}_{i,j} \in \mathbb{R}^{32 \times 32 \times C'} \). A pretrained 3D CNN $f_\theta: \mathbb{R}^{32 \times 32 \times C'} \rightarrow \mathbb{R}^{d'}$, as shown in (d), extracts spectral-spatial features $\mathbf{z}_{i,j} = f_\theta(\mathbf{P'}_{i,j})$, which are projected to dimension $d$ via a multilayer perceptron (MLP), producing aligned embeddings $\mathbf{z}'_{i,j} \in \mathbb{R}^d$. (c) Semantic alignment computes similarity scores $s^y_{i,j} = \mathbf{z}'_{i,j} \cdot \mathbf{e}_y$ for each class $y$, normalizes probabilities via softmax, \(p(y \mid \mathbf{P}'^i_j)\) and aggregates patch-level predictions through majority voting to determine the final class label for $\mathbf{X}_i$.
    }
    
    \label{fig:Overall_Architecture}
    
    \vspace{-4mm}
    
\end{figure*}

\noindent \textbf{Feature Extraction Models for HSI:}
2D CNNs primarily focus on extracting spatial features from hyperspectral images, treating them similarly to traditional RGB images. This approach uses a 2D convolutional kernel that slides across the spatial dimensions of the data \cite{simplified_2D_3DCNN_paper, Urban_Feature_Extraction_paper, HSIC_2d_3DCNN_paper}. Although effective for spatial analysis, 2D CNNs may ignore spectral relationships and correlations between different bands of a hyperspectral image. This can result in the loss of important spectral information that may improve the performance.
3D CNNs are designed to extract both spectral and spatial features from hyperspectral images simultaneously \cite{A_3D_Deep_CNN_paper}. It applies a 3D convolutional kernel that operates on all three dimensions of the hyperspectral data (spatial height, spatial width, and spectral bands). 3D CNNs can learn the interdependencies and correlations between adjacent spectral bands, allowing for a more comprehensive representation. However, 3D CNN-based networks lack an understanding of the semantic relationships between classes. In order to address these issues, our work focuses on using label semantics to guide the model.

\noindent \textbf{NLP methods on HSI:} Recent advances in HSI classification have explored the use of semantic representations to enhance generalization in diverse dataset. \cite{HSIC_ErtingPan} employ word2vec embeddings to bridge known and unknown categories through a three-phase scheme: feature embedding, mapping, and label reasoning. Similarly, LPFormer utilizes contrastive learning to align visual and textual characteristics \cite{LPFormer}, facilitating domain generalization. Other approaches integrate language models directly with hyperspectral data. Prompt-RSVQA extracts contextual information from visual data and processes it using DistilBERT to improve the accuracy of question answering \cite{RSVQA}. HSI-BERT combines BERT with hyperspectral images, offering robust generalization, spatial-spectral representation, and computational efficiency over CNNs \cite{HSI-BERT}. Furthermore, other BERT-based models improve HSI segmentation by incorporating spatial context via segment embeddings, outperforming traditional CNNs \cite{segment}. In cross-modal retrieval, KCR leverages knowledge graphs and sentence transformers to improve text-image matching \cite{knowledge}. Although these techniques effectively utilize semantic relationships and language models, they rely on predefined labels or embeddings. In this paper, we explore the potential of LLM to generate more contextually rich class descriptions to further improve HSI classification.

\begin{table*}[!t]

\centering

\caption{Example input prompt and LLM-generated semantic description for the {Heartwood} class. The texts are generated using OpenAI’s GPT-4 API and contrast {Heartwood} with {Sapwood} to enrich semantic embeddings for feature-label alignment.}

\vspace{-2mm}

\setlength{\tabcolsep}{3pt} 
\renewcommand{\arraystretch}{1.2}

\begin{tabular}{l|p{13.5cm}}
\hline
\text{Example Input Prompt} & Write a detailed factual paragraph about \textbf{[Heartwood]}, highlighting its \textbf{unique characteristics} and how it differs from the other labels: \textbf{[Sapwood]}. Focus on how they \textbf{look} different from each other and how their \textbf{reflectance} varies on the \textbf{NIR}, \textbf{VNIR} and \textbf{SWIR} spectral range. Also focus on \textbf{physical features } relevant to hyperspectral analysis. \\ \hline
\text{LLM(GPT-4) Generated Description} & \textbf{Heartwood} is the \textbf{dense}, \textbf{innermost} part of a tree trunk, typically \textbf{darker} in color compared to \textbf{Sapwood} due to the accumulation of extractives such as tannins, resins, and lignins. Spectrally, \textbf{Heartwood} and \textbf{Sapwood} exhibit distinct differences in their \textbf{reflectance profiles} across the \textbf{Near-Infrared (NIR)}, \textbf{Visible-Near Infrared (VNIR)}, and \textbf{Short-Wave Infrared (SWIR)} ranges. \textbf{Heartwood} generally has \textbf{lower reflectance} in the \textbf{VNIR} due to its \textbf{higher pigmentation} and \textbf{denser} structure, which leads to increased absorption of visible light. In the \textbf{NIR} range, \textbf{Heartwood} tends to reflect less than \textbf{Sapwood} because of its lower moisture.... \\ 
\hline

\end{tabular}
\label{tab:description_table}

\vspace{-5mm}

\end{table*}

\section{Method}

\subsection{Problem Formulation}
Assume a HSI dataset $\mathcal{D}$, where each HSI image is denoted as $\mathbf{X}_i$ with spatial dimensions $H \times W$ and $C$ spectral bands and is associated with a class label $y_i \in \mathcal{Y}$. The set of class labels is given by $\mathcal{Y} = \{y_1, \dots, y_U\}$, where $U$ denotes the number of unique classes. For each class $y \in \mathcal{Y}$, we associate a $d$-dimensional semantic label embedding $\mathbf{e}_y$ derived from an encoder. The set of text embeddings is represented as: ${\mathbf{E} = \{\mathbf{e}_y : y \in \mathcal{Y}\}, \quad \mathbf{e}_y \in \mathbb{R}^d.}$ The final dataset can be represented as ${\mathcal{D}^F = \{ (\mathbf{X}_i, y_i, \mathbf{e}_{y_i}) \mid i = 1, \dots, N \}}$, where $\mathbf{X}_i \in \mathbb{R}^{H \times W \times C}$ is the HSI image, $y_i \in \mathcal{Y}$ is the class label of $\mathbf{X}_i$, and $\mathbf{e}_{y_i} \in \mathbb{R}^{d}$ is the corresponding semantic label embedding. A function \( f_\theta \) is trained to extract feature embeddings \( \mathbf{z}_i \) from each HSI image, \(\mathbf{z}_i = f_\theta(\mathbf{X}_i), \mathbf{z}_i \in \mathbb{R}^d.\)

\noindent\textbf{Solution Strategy:} The goal is to predict the class label \(\hat{y}^i \in Y\) for each HSI image \( X^i \) by computing a probability distribution \( p(y \mid X^i) \) over all classes. This similarity score, \(s^y\), is derived from the alignment of spectral-spatial feature embeddings \( z_i \) and semantic label embeddings \( e_y \), ultimately selecting the class with the highest probability. This approach leverages both the spectral features of HSIs and semantic label embeddings to enhance classification performance.

Hyperspectral images (HSIs) are characterized by abundant spectral resolution, capturing detailed spectral-spatial information across numerous contiguous bands. Effectively utilizing this high-dimensional data presents challenges in feature extraction and classification. In Figure \ref{fig:Overall_Architecture} the architecture of our proposed \textbf{Semantic Spectral-Spatial Fusion Network (S3FN)}, a novel two-stage deep learning framework, is given. The model processes \(32 \times 32\) spectral-spatial cuboids, which are designed not only to optimize feature extraction given the high spectral resolution but also to ensure consistency in input dimensions. Additionally, this approach enables a voting mechanism during testing, where predictions from multiple patches of a single image can be aggregated for improved classification accuracy. By combining a 3D Convolutional Neural Network (3D CNN) for spectral-spatial feature extraction with semantic label embeddings for contextual understanding, the framework achieves robust and accurate classification of HSIs.

\subsection{LLM-generated Label Semantics for HSI Classification}
\label{subsec:LLM_Semantics}

Hyperspectral image (HSI) classification often faces challenges due to the limited availability of training samples and the high dimensionality of spectral data. To address this, we incorporate a LLM to generate enhanced semantic descriptions for class labels, which are then embedded into a vector space using a pre-trained text encoder to extract meaningful label semantics. These semantic embeddings are used to improve semantic feature alignment, enrich class semantics and provide contextual depth, enabling better classification performance .

Given a class label $y \in \mathcal{Y}$, we prompt the LLM with queries designed to generate descriptive paragraphs (approximately 10) that capture the specific characteristics of the class. 
These generated descriptions, denoted as $T_y$, provide a deeper understanding of the class semantics by emphasizing features aligned with the spectral characteristics. We define the prompts generated for each sample as: 

\[
\mathbf{T_y} = \mathbf{LLM}(\mathbf{Prompt_y}).
\] 

Examples of a class name description produced by the LLM(GPT-4) is provided in Table \ref{tab:description_table}.
We pass the generated semantic descriptions through a transformer-based text encoder, such as BERT or RoBERTa, to use this information. We extract the embedding corresponding to the class label token from the last transformer layer. This approach ensures that the final semantic representation $\mathbf{e}_y$ is directly influenced by the contextual information provided by the entire generated description.

The resulting semantic embeddings $\mathbf{e}_y$ are then aligned with the spectral characteristics extracted from the HSI patches using a 3D-CNN. By integrating contextual knowledge from the LLM with spectral features, we achieve improved feature-label alignment, improving the model's ability to identify unique spectral patterns associated with different classes.

Through this innovative integration of LLM-generated semantic descriptions, our method addresses key challenges in HSI classification, providing a powerful framework for using contextual knowledge and spectral information.

\subsection{Proposed Architecture}
\label{subsec:Proposed_Architecture}

For \(\mathcal{D^F}\) which is the HSI dataset, where each image \(\mathbf{X}_i \in \mathbb{R}^{H \times W \times C}\) has spatial dimensions \(H \times W\) and \(C\) spectral bands. Each image is divided into \(M\) non-overlapping patches, \(\mathbf{P}_i = \{\mathbf{P}_{i,j} \in \mathbb{R}^{32 \times 32 \times C} \mid j = 1, \dots, M\}\). The patched dataset is represented as $\mathcal{P} = \left\{ \left( \{\mathbf{P}_{i,j}\}_{j=1}^{M}, y_i \right) \mid i = 1, \dots, N \right\}$, where $\mathbf{P}_{i,j} \in \mathbb{R}^{32 \times 32 \times C}$ is the \( j \)-th patch of image \( \mathbf{X}_i \).To address the high dimensionality of hyperspectral data and enhance computational efficiency, we incorporated PCA as a preprocessing step to reduce the spectral dimensions of the input patches. Given a hyperspectral image \( X_i \in \mathbb{R}^{H \times W \times C} \), each non-overlapping patch \( P_{i,j} \in \mathbb{R}^{32 \times 32 \times C} \) was reshaped into a 2D matrix \( P_{\text{flat}} \in \mathbb{R}^{(32 \times 32) \times C} \), where each row represents the spectral signature of a spatial pixel. PCA was applied globally across all training patches to compute a transformation matrix that projects the original \( C \)-dimensional spectral vectors into a reduced \( C' \)-dimensional subspace (\( C' \ll C \)), preserving 99\% of the cumulative variance. This resulted in compressed patches \(\mathbf{P'}_{i,j} \in \mathbb{R}^{32 \times 32 \times C'} \), which were then fed into the 3D-CNN.

\begin{figure*}[!t]

    \centering
    
    \includegraphics[width=1\linewidth]{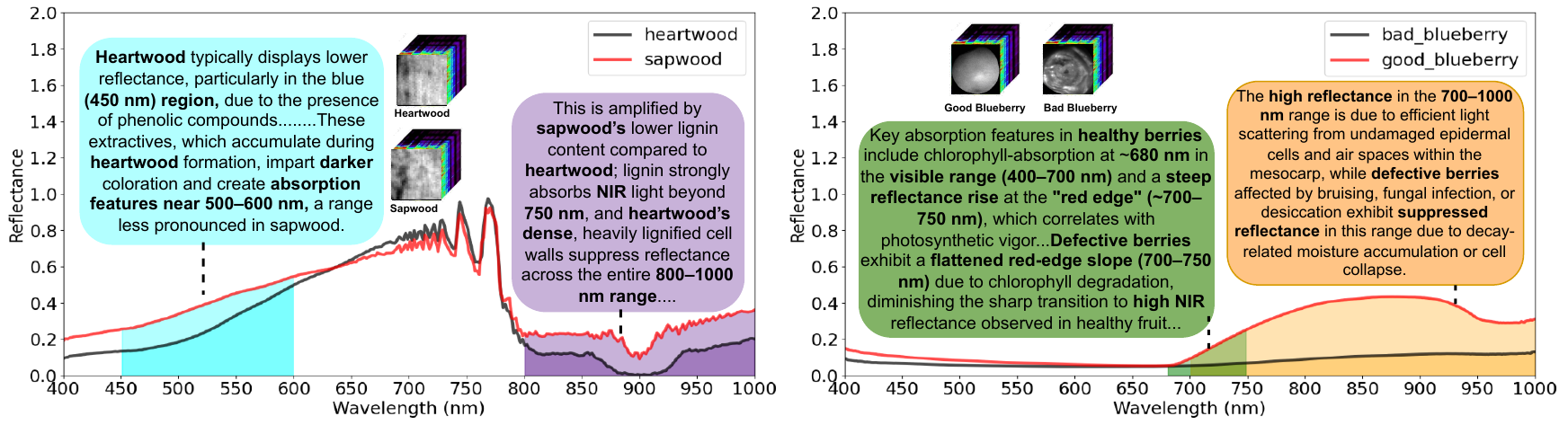}
    
     \caption{Mean spectral reflectance curves for two unique samples of wood (left: heartwood and sapwood) and blueberries (right: healthy/good and bad/defective), illustrating key absorption and reflectance features. The spectral mean reflectance curve for each sample was computed as described in Section \ref{subsec:Proposed_Architecture}, averaging reflectance values across all pixels for each spectral band. The overlaid semantic descriptions, generated by a LLM, capture class-specific spectral characteristics (highlighted in color-coded regions), enhancing label embeddings for robust hyperspectral image classification. 
     For example, \textbf{Heartwood} typically displays \textbf{lower reflectance}, particularly in the \textbf{blue (450 nm) region}, due to the presence of phenolic compounds.
     }

 \label{fig:wood_bb_spectral2}
 
 \vspace{-5mm}
 
\end{figure*}

\noindent\textbf{Stage 1. Spectral-Spatial Feature Extraction with 3D-CNN:}  
3D Convolutional Neural Network (3D-CNN) extracts spectral-spatial features from each patch \(\mathbf{P'}_{i,j}\). Let the output of the 3D-CNN for a patch \(\mathbf{P"}_{i,j}\) be denoted as \(\mathbf{z}_{i,j} \in \mathbb{R}^{d}\), representing its feature embedding:

\vspace{-3.5mm}

\[
\mathbf{z}_{i,j} = f_\theta(\mathbf{P'}_{i,j}),
\]
where \(f_\theta\) is the 3D-CNN parameterized by \(\theta\).

\noindent The 3D-CNN consists of multiple convolutional layers. At the \(l\)-th layer, the feature maps are updated as:
\[
\mathbf{x}_{i,j}^{(l+1)} = \sigma\left(\mathbf{W}^{(l)}* \mathbf{x}_{i,j}^{(l)} + \mathbf{b}^{(l)}\right),
\]

where \(\mathbf{x}_{i,j}^{(l)}\) represents the input at layer \(l\), \(*\) denotes the 3D convolution operation with kernel size \(3 \times 3 \times 3\), \(\sigma\) is the ReLU activation function, and \(\mathbf{W}^{(l)}\) and \(\mathbf{b}^{(l)}\) are the trainable weights and biases.
To prevent overfitting, L2 regularization is applied to the weights:

\vspace{-2mm}

\[
\mathcal{L}_{\text{reg}} = \frac{\lambda}{2} \sum_{l} \|\mathbf{W}^{(l)}\|_2^2,
\]
where \(\lambda\) is the regularization coefficient. Average pooling layers progressively reduce spatial and spectral dimensions:
\[
\mathbf{x}_{i,j}^{(l+1)} = \text{AvgPool}\left(\mathbf{x}_{i,j}^{(l)}, k\right),
\]
where \(k\) is the pooling window size.

\noindent After the final convolutional layer, the feature maps are flattened and passed through two fully connected layers with dropout:

\vspace{-6mm}

\[
\mathbf{f}_{i,j} = \text{Dropout}\left(\sigma\left(\mathbf{W}_{\text{fc1}} \cdot \mathbf{v}_{i,j} + \mathbf{b}_{\text{fc1}}\right)\right),
\]

\vspace{-5mm}

\[
\mathbf{z}_{i,j} = \sigma\left(\mathbf{W}_{\text{fc2}} \cdot \mathbf{f}_{i,j} + \mathbf{b}_{\text{fc2}}\right),
\]
where \(\mathbf{v}_{i,j}\) is the flattened output of the final pooling layer, and \(\mathbf{W}_{\text{fc1}}, \mathbf{W}_{\text{fc2}}, \mathbf{b}_{\text{fc1}}, \mathbf{b}_{\text{fc2}}\) are trainable parameters.

\noindent\textbf{Stage 2. Semantic Alignment with Label Embeddings:}  
Each class \( y \in \mathcal{Y} \) is associated with a semantic class embedding \( \mathbf{e}_y \in \mathbb{R}^d \), forming a set \( \mathbf{E} = \{\mathbf{e}_y : y \in \mathcal{Y}\} \). The embeddings are generated using the LLM generated class descriptions as textual guidance for the encoders. The image feature \( \mathbf{z}_{ij} \) is then passed through an MLP resulting in \( \mathbf{z}'_{ij} \).

For a patch \(\mathbf{P}_{i,j}\), the similarity score with a class \(y\) is computed as:

\vspace{-5mm}

\[
s_{i,j}^y = \mathbf{z'}_{i,j}^\top . \mathbf{e}_y,
\]

\vspace{-1mm}

where \(s_{i,j}^y\) quantifies the alignment between the spectral-spatial features \(\mathbf{z'}_{i,j}\) and the semantic embedding \(\mathbf{e}_y\).
Class probabilities are obtained by normalizing the similarity scores using a softmax function:

\vspace{-3mm}

\[
p(y \mid \mathbf{P'}_{i,j}) = \frac{\exp(s_{i,j}^y)}{\sum_{y' \in \mathcal{Y}} \exp(s_{i,j}^{y'})}.
\]

At inference, each patch independently predicts a class label, and the final class for the entire image for \(X^i\) is determined through majority voting among the patch-level predictions.

\noindent\textbf{Loss Function:}
The overall loss function includes classification loss and regularization:

\vspace{-4mm}

\[
\mathcal{L} = \mathcal{L}_{\text{CE}} + \mathcal{L}_{\text{reg}},
\]

\vspace{-2mm}

where the classification loss \(\mathcal{L}_{\text{CE}}\) is defined as:
\[
\mathcal{L}_{\text{CE}} = - \frac{1}{N} \sum_{i=1}^N \sum_{j=1}^M \log\left(p(y_i \mid \mathbf{P'}_{i,j})\right).
\]

By integrating spectral-spatial features from \(\mathbf{z}_{i,j}\) with semantic embeddings \(\mathbf{e}_y\), the proposed architecture leverages both hyperspectral characteristics and label semantics to improve classification accuracy.

\begin{table*}[!t]

\centering

\caption{Detailed statistics of Datasets used.}

\vspace{-2mm}

\setlength{\tabcolsep}{9pt}
\renewcommand{\arraystretch}{0.5}
\begin{tabular}{ccccccccc}
\toprule
\textbf{Dataset} & \textbf{Resolution} & \textbf{Spectral}  & \textbf{Spectral} & \textbf{Sample} & \textbf{Class} & \multicolumn{3}{c}{\textbf{Dataset Split(\%)}} \\
\cmidrule(r){7-9}
 \textbf{Name} &   &\textbf{Range (nm)} & \textbf{Bands} & \textbf{Size} & \textbf{no.} & \textbf{Train} &\textbf{Test} &\textbf{Val.} \\
\midrule
Hyperspectral Wood \cite{wood_dataset}
&32 × 32 × 320     &400-1000  &320  &108 &2 &67 &33 &-- \\
\midrule
HyperspectralBlueberries \cite{blueberry_dataset}  &1800 × 1600 
× 462  &400-1000  &447 &462 &2 &70 &30 &-- \\
\midrule
DeepHS-Fruit \cite{ripeness_paper_dataset} & 400 × (250-350) × (224-252)   &400-1700  &224-252  &442 & 3 &75  &12.5 &12.5 \\
\bottomrule
\end{tabular}
\label{tab:dataset-details}

\vspace{-6mm}

\end{table*}

In Fig. \ref{fig:wood_bb_spectral2}, we demonstrate how LLM-generated class descriptions truly align with the mean spectral reflectance curves of two random samples from the Hyperspectral Wood \cite{wood_dataset} and HyperspectralBlueberries \cite{blueberry_dataset} datasets. The spectral mean reflectance for each sample was computed by averaging the reflectance values across all pixels for each spectral band. For a HSI, \( X^i\), of dimensions \( H \times W \times C \), where \( H \) and \( W \) represent the spatial dimensions, and \( C \) denotes the number of spectral bands. For a given band \( c \), the mean spectral reflectance \( \bar{R}_c \) is calculated as: \(\bar{R}_c = \frac{1}{H W} \sum_{i=1}^{H} \sum_{j=1}^{W} R_{i,j,c}\), where \( R_{i,j,c} \) represents the reflectance value at pixel location \( (i,j) \) for the spectral band \( c \). The complete spectral mean curve for the sample is then represented as the vector: \(\bar{R} = (\bar{R}_1, \bar{R}_2, \dots, \bar{R}_C).\). This comparison between the LLM-generated descriptions and the mean spectral curves validates that the descriptions contain accurate spectral distinctions (e.g. highlighting absorption features and differentiating between healthy vs. defective blueberries or heartwood vs. sapwood).

\section{Experiments}

\subsection{Setup}
\label{subsec:Experiments_Setup}

\noindent\textbf{Datasets:} We have used three well-known HSI datasets:
\textbf{(1)} Hyperspectral Images for Wood Recognition \cite{wood_dataset},
\textbf{(2)} HyperspectralBlueberries \cite{blueberry_dataset} dataset to identify good and defective blueberries and \textbf{(3)} DeepHS-Fruit \cite{ripeness_paper_dataset} to determine the ripeness state of fruits (Avocado and Kiwi).
Table \ref{tab:dataset-details} illustrates further details about the used datasets.

\noindent\textbf{Evaluation metric:} Following previous works \cite{potato_paper, orange_paper, blueberry_paper, paddy_seed}, we use accuracy, precision, recall, and the F1 score to evaluate different HSI classification methods.

\noindent\textbf{Implementation details:} 3D-CNN comprises three convolutional layers with ${3 \times 3 \times 3}$ kernels, each with ReLu activation and a progressively increasing number of filters (32, 64, 128). The average pooling layers follow the first and third convolutional layers. The flattening layer is followed by two FC layers with 128 and 64 units, using ReLU activations and L2 regularization; a dropout rate of 0.5 is applied after the first dense layer. The output layer is a dense layer with two units and a softmax activation.
\textit{Semantic Unit:} Class descriptions were generated using an LLM through OpenAI’s GPT-4 API and encoded using encoders such as BERT, RoBERTa, and Word2Vec.
\textit{Fusion Mechanism:} The 3DCNN's layers are frozen to retain previously learned features; the output from the FC layer after the flattening layer is used as input for the fusion process.
A new dense layer or MLP with ReLU activation is added to learn more complex mappings between the image features and the label embeddings and to project the image features to the same dimension as the label embeddings.
The dot product is computed between the logits from the MLP and the Label Embeddings, creating a similarity matrix that captures relationships between the input features and semantic embeddings.
The similarity matrix is passed through a softmax activation layer to produce probabilistic predictions.
\textit{Training Configuration:} Random scaling between 0.9 and 1.1 was applied as a data augmentation. Sparse categorical cross-entropy loss and Adam Optimizer, a batch size of 4, are used to train for 100 epochs. The L2 regularization of ${1 \times 10^{-2}}$ is applied to the convolutional and FC layers to avoid overfitting. The model was implemented using an NVIDIA RTX 3060 GPU using the TensorFlow framework.

\subsection{Main Results}

In Table \ref{tab:comparison}, the result of our proposed model is compared to Cifar10Net\cite{wood_paper}, RDLA \& LDA \cite{blueberry_paper}, and HS\_CNN \cite{ripeness_paper_dataset}, in their respective datasets. Our method continuously beats the other models in every important metric, proving its resilience and efficiency for the tasks at hand. The following table \ref{tab:comparison} provides a comprehensive evaluation of precision metrics (PR), recall, F1 score, and accuracy (Acc).

\begin{table}[!t]

\centering

\caption{Performance Comparison of different HSI datasets. \textbf{RoBERTa} is used as a text encoder for all experiments.}

\vspace{-2mm}

\setlength{\tabcolsep}{13pt}
\renewcommand{\arraystretch}{1}
\begin{tabular}{@{}lc|c|c|c@{}}
    \toprule
    \multicolumn{5}{c}{\textbf{Hyperspectral Wood \cite{wood_dataset}}} \\
    \midrule
    \textbf{Model} & \textbf{PR} & \textbf{Recall} & \textbf{F1} & \textbf{ACC} \\
    \midrule
    SVM     & 89.0 & 89.0 & 89.0 & 88.6 \\
    KNN     & 84.0 & 84.0 & 84.0 & 84.0 \\
    Random Forest      & 82.0 & 82.0 & 82.0 & 81.1 \\
    Neural Network      & 88.0 & 87.0 & 87.0 & 87.1 \\
    Decision Tree     & 59.0 & 58.0 & 58.0 & 58.3 \\
    Cifar10Net\cite{wood_paper} & - & - & - & 93.9 \\
    S3FN (Ours)    & \textbf{95.0} &\textbf{95.0} &\textbf{95.0} & \textbf{94.7} \\
    \midrule
     \multicolumn{5}{c}{\textbf{HyperspectralBlueberries \cite{blueberry_dataset}}} \\
    \midrule
    \textbf{Model} & \textbf{PR} & \textbf{Recall} & \textbf{F1} & \textbf{ACC} \\
    \midrule
    SVM     & 92.0 & 92.0 & 92.0 & 91.7 \\
    KNN     & 76.0 & 75.0 & 75.0 & 75.2 \\
    Random forest      & 77.0 & 76.0 & 77.0 & 76.4 \\
    Neural Network      & 86.0 & 86.0 & 86.0 & 85.8 \\
    Decision Tree     & 81.0 & 80.0 & 80.0 & 80.0 \\
    LDA  \cite{blueberry_paper}          &90.8 &78.6 & - &85.3\\
    RLDA  \cite{blueberry_paper}         &\textbf{97.7} &93.7 & - &95.7\\
    RLDA\&LDA \cite{blueberry_paper}  & 96.5 & \textbf{96.7} & - & \textbf{96.6} \\
    S3FN (Ours)    & 86.0 & 86.0 & 86.0 & 86.4 \\
\end{tabular}
\label{tab:comparison}

\setlength{\tabcolsep}{11pt}
\renewcommand{\arraystretch}{1}
\begin{tabular}{@{}lcc|cc@{}}
    \toprule
     \multicolumn{5}{c}{\textbf{DeepHS-Fruit \cite{ripeness_paper_dataset}}} \\
    \midrule
    \textbf{Model} & \multicolumn{2}{c|}{\textbf{Ripeness (C1)}} & \multicolumn{2}{c}{\textbf{Ripeness (C2)}} \\
    \cmidrule(lr){2-3} \cmidrule(lr){4-5}
    & \textbf{Avocado} & \textbf{Kiwi} & \textbf{Avocado} & \textbf{Kiwi} \\
    \midrule
    SVM  & 57.1  &55.5 &80.0 &56.5\\
    KNN  & 57.1  &33.3 &86.6 &65.2\\
    Random Forest  &53.3 &57.8 &87.0& 61.7 \\
    Neural Network   & 80.0  &78.9  
    & 93.5 & 76.5 \\
    Decision Tree   & 80.0  & 42.1  & 70.9 & 53.1\\
    ResNet-18\cite{ripeness_paper_dataset} & 44.4 & 60.0  & 66.7 & 33.3 \\
     AlexNet \cite{ripeness_paper_dataset} & 33.3 & 33.3  & 33.3 & 33.3 \\
    HS-CNN \cite{ripeness_paper_dataset} & 44.4 & 66.7  & 33.3 & 33.3 \\
    S3FN (Ours) & \textbf{66.7} & \textbf{70.4} & \textbf{47.1} & \textbf{44.8}\\
    \bottomrule
\end{tabular}

\vspace{-6mm}

\end{table}

\noindent\textbf{Hyperspectral Wood \cite{wood_dataset} :} In Table \ref{tab:comparison}, our proposed model achieves the highest precision, \textbf{94. 5\%}, surpassing all baseline methods. This illustrates a significant improvement over the Cifar10Net~\cite{wood_paper} model of the Hyperspectral Wood dataset, which achieves 93.9\%. Among the baseline models, the SVM performs best with an accuracy of 88.6\%, whereas other methods do not. This highlights our model's capability to capture spatial-spectral features via the 3D-CNN and that LLM-generated semantic embeddings of class labels provide a deeper contextual understanding of class characteristics and allow for better feature-label alignment. This is opposed to Cifat10Net's modified 2D-CNN approach, which does not capture spectral relationships as effectively as a 3D-CNN but does have a lower computational cost.

\begin{table*}[!t]
\centering
\caption{Performance using different text encoder for semantic embeddings. We obtain our best results using \textbf{RoBERTa}.}

\vspace{-2mm}

\setlength{\tabcolsep}{1.5pt}
\renewcommand{\arraystretch}{0.5}

\begin{tabular}{@{}lccc|ccc|ccc|ccc@{}}
    \toprule
    \textbf{Dataset} & \multicolumn{3}{c|}{\textbf{PR (\%)}} & \multicolumn{3}{c|}{\textbf{Recall (\%)}} & \multicolumn{3}{c|}{\textbf{F1(\%)}} & \multicolumn{3}{c}{\textbf{ACC (\%)}} \\
    \cmidrule(lr){2-4} \cmidrule(lr){5-7} \cmidrule(lr){8-10} \cmidrule(lr){11-13}
    & \textbf{BERT} & \textbf{RoBERTa} & \textbf{Word2Vec} 
    & \textbf{BERT} & \textbf{RoBERTa} & \textbf{Word2Vec} 
    & \textbf{BERT} & \textbf{RoBERTa} & \textbf{Word2Vec} 
    & \textbf{BERT} & \textbf{RoBERTa} & \textbf{Word2Vec} \\
    \midrule
    Hyperspectral Wood \cite{wood_dataset} & 93.2 & \textbf{95.0} & 92.0 & 93.2 & \textbf{95.0}& 92.0 & 93.2& \textbf{95.0}& 92.0 & 93.2& \textbf{94.7}& 91.8\\
    DeepHS-Fruit (C1 Avocado) \cite{ripeness_paper_dataset} &  51.5& \textbf{53.9}  & 47.0& 64.0& \textbf{66.7} & 60.5& 57.1& \textbf{57.2} & 52.9& 65.5& \textbf{66.7} & 61.8\\
    \bottomrule
\end{tabular}
\label{tab:performance_comparison}

\vspace{-6mm}

\end{table*}

\noindent\textbf{HyperspectralBlueberries\cite{blueberry_dataset} :} The RLDA \& LDA \cite{blueberry_paper} in Table \ref{tab:comparison} provides the best performance with an accuracy of \textbf{96.6\%}. Our proposed model was unable to outperform them. This highlights the remarkable proficiency of classical machine learning models in the HSI domain, with SVM and KNN achieving respectable performance throughout the paper across all three datasets. Although deep learning techniques have gained popularity, classical methods like Linear Discriminant Analysis (LDA), Regularized Linear Discriminant Analysis (RLDA), and Support Vector Machines(SVM) can achieve high accuracy with relatively less computational cost.

An important factor to consider is that, unlike the other two datasets, Hyperspectral Wood \cite{wood_dataset} and DeepHS-Fruit \cite{ripeness_paper_dataset}, the HyperspectralBlueberries \cite{blueberry_dataset} dataset did not include a preprocessed version. To ensure a fair comparison, we tried to replicate the pre-processing steps as accurately as possible, following the image processing steps provided by Deng et al.\cite{blueberry_paper}. However, slight differences in preprocessing may have contributed to the performance discrepancy observed between our model, S3FN, and models such as RLDA and LDA\cite{blueberry_paper}. Furthermore, we used PCA-reduced data instead of full hyperspectral images because of computational resource constraints, which may have further influenced model performance.

\noindent\textbf{DeepHS-Fruit \cite{ripeness_paper_dataset}:} Table \ref{tab:comparison} compares the model performance on the DeepHS-Fruit dataset, which includes hyperspectral data from two camera systems (C1: INNO-SPEC RedEye; C2: Specim FX10) for avocado and kiwi ripeness classification. For \textbf{C1 (Ripeness)}, our model (S3FN) achieves an accuracy of \textbf{66.7\%} on avocado, outperforming ResNet-18 (44.4\%), AlexNet (33.3\%) and HS-CNN (44.4\%) \cite{ripeness_paper_dataset}. In particular, we compared the PCA-based performance of HS-CNN with our PCA-integrated S3FN model, highlighting that S3FN utilizes dimensionally reduced data from PCA, while providing superior performance. However, the neural network and decision tree baseline methods achieve higher performance at 80.0\%. For \textbf{kiwi under C1}, S3FN attains the highest accuracy (\textbf{70.4\%}), surpassing HS-CNN (66.7\%) and all other baselines except Neural Network (78.9\%).For \textbf{C2 (Ripeness)}, S3FN achieves \textbf{50.0\%} on avocado and \textbf{44.8\%} on kiwi, exceeding HS-CNN’s performance (33.3\% for both fruits). Although the neural network remains the strongest baseline for the C2 avocado (93. 5\%), S3FN outperforms HS-CNN by significant margins. Other baselines such as KNN (86. 6\%) and Random Forest (87.0\%) also show competitive results for the C2 avocado. The superiority of our model over HS-CNN is due to its integration of \textbf{LLM-generated semantic embeddings} to enrich hyperspectral feature representation, allowing better alignment of the characteristic label. Unlike HS-CNN, which relies solely on spectral-spatial features through depth-wise separable convolutions, S3FN employs a \textbf{3D-CNN} for robust spectral-spatial feature extraction combined with semantic context. This hybrid approach enhances discriminative power, particularly in complex ripeness classification tasks.

\subsection{Ablation Studies}

\noindent\textbf{Different text encoders:} In Table \ref{tab:performance_comparison}, we compare different text encoders (BERT \cite{BERT}, RoBERTa \cite{RoBERTa}, and Word2Vec \cite{W2V}) to analyze their ability to generate semantic embeddings from class descriptions, revealing that transformer-based encoders produce superior representations. For transformer-based models, we extracted embeddings from the final transformer layer after processing the entire corpus, allowing rich contextualized representations. In contrast, for Word2Vec, we computed the average of all word vectors from the corpus to obtain a single representation, which relies on static embeddings and lacks contextual nuance. These findings suggest that contextualized embeddings capture richer semantic information, enabling better representation of class descriptions. Among transformer-based models, RoBERTa slightly outperforms BERT, probably due to its improved pre-training strategies, such as larger training data and dynamic masking \cite{RoBERTa}. In contrast, Word2Vec, which relies on static word embeddings \cite{W2V}, struggles to capture nuanced contextual meanings, leading to lower performance.

\begin{table}[!t]
\centering

\caption{ 
Using different network configurations (RoBERTa as Text encoder) on Hyperspectral
Wood\cite{wood_dataset} and DeepHS-Fruit\cite{ripeness_paper_dataset} datasets}

\vspace{-2mm}

\setlength{\tabcolsep}{3pt}
\renewcommand{\arraystretch}{1}
\begin{tabular}{@{}lc|cc|cc@{}}
    \toprule
    \multicolumn{2}{c|}{\textbf{Hyperspectral Wood \cite{wood_dataset}}} & \multicolumn{4}{c}{\textbf{DeepHS-Fruit\cite{ripeness_paper_dataset}}} \\
    \cmidrule(lr){1-2} \cmidrule(lr){3-6}
    \textbf{Configuration}  & \textbf{Acc.}  & \multicolumn{2}{c|}{\textbf{Ripeness C1 (Acc.)}} & \multicolumn{2}{c}{\textbf{Ripeness C2 (Acc.)}} \\
    \cmidrule(lr){3-4} \cmidrule(lr){5-6}
    & & \textbf{Avocado} & \textbf{Kiwi} & \textbf{Avocado} & \textbf{Kiwi} \\
    \midrule
     Standalone 3D CNN  & 88.0  & 55.6 & 66.7 & 47.0 & 43.0 \\
     No description     & 93.9  & 57.5 & 68.0 & 47.0 & 44.5 \\
     S3FN (Ours)        & \textbf{94.7}  &  \textbf{66.7} & \textbf{70.4} & \textbf{47.1} & \textbf{44.8} \\  
    \bottomrule
\end{tabular}
\label{tab:Freezing_LayerSize}

\vspace{-6mm}

\end{table}

\noindent\textbf{Different network configurations:} 
Table \ref{tab:Freezing_LayerSize} examines the contribution of text embeddings by comparing the embeddings of the labels alone versus our model, which integrates label embeddings with LLM-generated descriptions, using a standalone 3D-CNN as a baseline. The results demonstrate that our model outperforms both alternatives. 

\subsection{Limitations and Future Work}

While this work introduces a semantic alignment framework for HSI classification, several limitations remain. First, using GPT-4 directly to classify hyperspectral images is not possible. These images contain hundreds of spectral bands, but GPT-4 can only handle text or low-resolution images. Turning hyperspectral data into a format GPT-4 can read would remove important spectral details. Second, models like VLMs (e.g., CLIP) are trained on regular RGB images, not hyperspectral data. They do not understand the extra spectral information and can not be used directly without significant retraining, which is not currently available. Our method avoids this by creating semantic features matching hyperspectral images' unique properties. In future work, we plan to test how the model performs if the text descriptions are noisy or wrong and explore ways to train models that learn from both text and hyperspectral data together without needing hand-written prompts.

\section{Conclusion}

This paper explores the classification of HSI through the integration of LLM-guided semantics. We propose a Semantic Spectral-Spatial Fusion Network (S3FN), a novel framework that utilizes an LLM to generate rich, contextually detailed textual descriptions for each class label, capturing their unique spectral characteristics and differences. These descriptions are transformed into semantic embeddings via text encoders, enriching the contextual representation of class labels. Our model improves the alignment between label semantics and spectral features by aligning these semantic embeddings with spectral-spatial features extracted through a 3D-CNN. This integration of linguistic and spectral insights paves the way for more robust and generalizable approaches to HSI classification across diverse application domains.

\bibliographystyle{IEEEtran}
\bibliography{ref}

\begin{thebibliography}{10}
\providecommand{\url}[1]{#1}
\csname url@samestyle\endcsname
\providecommand{\newblock}{\relax}
\providecommand{\bibinfo}[2]{#2}
\providecommand{\BIBentrySTDinterwordspacing}{\spaceskip=0pt\relax}
\providecommand{\BIBentryALTinterwordstretchfactor}{4}
\providecommand{\BIBentryALTinterwordspacing}{\spaceskip=\fontdimen2\font plus
\BIBentryALTinterwordstretchfactor\fontdimen3\font minus
  \fontdimen4\font\relax}
\providecommand{\BIBforeignlanguage}[2]{{%
\expandafter\ifx\csname l@#1\endcsname\relax
\typeout{** WARNING: IEEEtran.bst: No hyphenation pattern has been}%
\typeout{** loaded for the language `#1'. Using the pattern for}%
\typeout{** the default language instead.}%
\else
\language=\csname l@#1\endcsname
\fi
#2}}
\providecommand{\BIBdecl}{\relax}
\BIBdecl

\bibitem{simplified_2D_3DCNN_paper}
C.~Yu, R.~Han, M.~Song, C.~Liu, and C.-I. Chang, ``A simplified 2d-3d cnn
  architecture for hyperspectral image classification based on
  spatial–spectral fusion,'' \emph{IEEE Journal of Selected Topics in Applied
  Earth Observations and Remote Sensing}, vol.~13, pp. 2485--2501, 2020.

\bibitem{Urban_Feature_Extraction_paper}
X.~Ma, Q.~Man, X.~Yang, P.~Dong, Z.~Yang, J.~Wu, and C.~Liu, ``Urban feature
  extraction within a complex urban area with an improved 3d-cnn using airborne
  hyperspectral data,'' \emph{Remote Sensing}, vol.~15, no.~4, 2023.

\bibitem{HSIC_2d_3DCNN_paper}
Z.~Ge, G.~Cao, X.~Li, and P.~Fu, ``Hyperspectral image classification method
  based on 2d–3d cnn and multibranch feature fusion,'' \emph{IEEE Journal of
  Selected Topics in Applied Earth Observations and Remote Sensing}, vol.~13,
  pp. 5776--5788, 2020.

\bibitem{orange_paper}
R.~Pourdarbani, S.~Sabzi, R.~Zohrabi, G.~García-Mateos, R.~Fernandez-Beltran,
  J.~Molina~Martínez, and M.~H. Rohban, ``Comparison of 2d and 3d
  convolutional neural networks in hyperspectral image analysis of fruits
  applied to orange bruise detection,'' \emph{Journal of Food Science},
  vol.~88, 10 2023.

\bibitem{HSIC_ErtingPan}
E.~Pan, Y.~Ma, F.~Fan, X.~Mei, and J.~Huang, ``Hyperspectral image
  classification across different datasets: A generalization to unseen
  categories,'' \emph{Remote Sensing}, vol.~13, no.~9, 2021.

\bibitem{LPFormer}
B.~He, J.~Feng, D.~Li, R.~Shang, J.~Wu, and L.~Jiao, ``Learnable prompts-based
  transformers for domain generalization of hyperspectral image
  classification,'' in \emph{IGARSS 2024 - 2024 IEEE International Geoscience
  and Remote Sensing Symposium}, 2024, pp. 10\,441--10\,445.

\bibitem{wood_dataset}
{Roberto Confalonieri}, ``Hyperspectral images for wood recognition (sapwood
  and heartwood),'' 2022.

\bibitem{blueberry_dataset}
{Lu, Yuzhen}, ``Hyperspectralblueberries: a dataset of hyperspectral
  reflectance images of normal and defective blueberries,'' 2024.

\bibitem{ripeness_paper_dataset}
L.~A. Varga, J.~Makowski, and A.~Zell, ``Measuring the ripeness of fruit with
  hyperspectral imaging and deep learning,'' in \emph{2021 International Joint
  Conference on Neural Networks (IJCNN)}, 2021, pp. 1--8.

\bibitem{yang2024rapid}
W.~Yang, C.~Liu, S.~Zeng, X.~Duan, C.~Zhang, and W.~Tao, ``Rapid identification
  of moldy peanuts based on three-dimensional hyperspectral object detection,''
  \emph{Journal of Food Composition and Analysis}, p. 106400, 2024.

\bibitem{huang2024detection}
H.~Huang, Y.~Liu, S.~Zhu, C.~Feng, S.~Zhang, L.~Shi, T.~Sun, and C.~Liu,
  ``Detection of mechanical damage in corn seeds using hyperspectral imaging
  and the resnest\_e deep learning network,'' \emph{Agriculture}, vol.~14,
  no.~10, p. 1780, 2024.

\bibitem{ozdougan2025wheat}
G.~{\"O}zdo{\u{g}}an and A.~Gowen, ``Wheat grain classification using
  hyperspectral imaging: Concatenating vis-nir and swir data for single and
  bulk grains,'' \emph{Food Control}, vol. 168, p. 110953, 2025.

\bibitem{zhao2024early}
X.~Zhao, Y.~Liu, Z.~Huang, G.~Li, Z.~Zhang, X.~He, H.~Du, M.~Wang, and Z.~Li,
  ``Early diagnosis of cladosporium fulvum in greenhouse tomato plants based on
  visible/near-infrared (vis/nir) and near-infrared (nir) data fusion,''
  \emph{Scientific Reports}, vol.~14, no.~1, p. 20176, 2024.

\bibitem{tao2024raman}
F.~Tao, H.~Yao, Z.~Hruska, K.~Rajasekaran, J.~Qin, M.~Kim, and K.~Chao, ``Raman
  hyperspectral imaging as a potential tool for rapid and nondestructive
  identification of aflatoxin contamination in corn kernels,'' \emph{Journal of
  Food Protection}, vol.~87, no.~9, p. 100335, 2024.

\bibitem{zhao2024multilayer}
Y.~Zhao, H.~Liu, X.~Zhai, R.~Zhang, W.~Shi, L.~Zhao, and Z.~Han, ``Multilayer
  spatial-spectral segmentation network for detecting afb1,'' \emph{Journal of
  Food Composition and Analysis}, vol. 136, p. 106790, 2024.

\bibitem{zhai2025simultaneously}
Y.~Zhai, J.~Wang, L.~Zhou, X.~Zhang, Y.~Ren, H.~Qi, and C.~Zhang,
  ``Simultaneously predicting spad and water content in rice leaves using
  hyperspectral imaging with deep multi-task regression and transfer component
  analysis,'' \emph{Journal of the Science of Food and Agriculture}, vol. 105,
  no.~1, pp. 554--568, 2025.

\bibitem{li2024detection}
B.~Li, C.-t. Su, H.~Yin, J.-p. Zou, and Y.-d. Liu, ``Detection of moisture
  content of edamame based on the fusion of reflectance and transmittance
  spectra of hyperspectral imaging,'' \emph{Journal of Chemometrics}, vol.~38,
  no.~9, p. e3574, 2024.

\bibitem{ronay2024weed}
I.~Ronay, R.~N. Lati, and F.~Kizel, ``Weed species identification: Acquisition,
  feature analysis, and evaluation of a hyperspectral and rgb dataset with
  labeled data,'' \emph{Remote Sensing}, vol.~16, p. 2808, 2024.

\bibitem{wonggasem2024utilization}
K.~Wonggasem, P.~Wongchaisuwat, P.~Chakranon, and D.~Onwimol, ``Utilization of
  machine learning and hyperspectral imaging technologies for classifying
  coated maize seed vigor: A case study on the assessment of seed dna repair
  capability,'' \emph{Agronomy}, vol.~14, no.~9, p. 1991, 2024.

\bibitem{zhong2024intelligent}
Y.~Zhong, W.~Wen, X.~Fan, and N.~Cheng, ``An intelligent process analysis
  method for rapidly evaluating the quality of chinese medicine with
  near-infrared non-contact hyperspectral imaging: A case study of weifuchun
  concentrate,'' \emph{Phytochemical Analysis}, vol.~35, no.~7, pp. 1649--1658,
  2024.

\bibitem{xue2024rapid}
X.~Xue, H.~Tian, K.~Zhao, Y.~Yu, Z.~Xiao, C.~Zhuo, and J.~Sun, ``Rapid lactic
  acid content detection in secondary fermentation of maize silage using
  colorimetric sensor array combined with hyperspectral imaging,''
  \emph{Agriculture}, vol.~14, no.~9, p. 1653, 2024.

\bibitem{bai2024hyperspectral}
X.~Bai, Y.~You, H.~Wang, D.~Zhao, J.~Wang, and W.~Zhang, ``Hyperspectral
  reflectance imaging for visualizing reducing sugar content, moisture, and
  hollow rate in red ginseng,'' \emph{Heliyon}, vol.~10, no.~18, 2024.

\bibitem{rs16173230}
W.~Guo, X.~Xu, X.~Xu, S.~Gao, and Z.~Wu, ``Clustering hyperspectral imagery via
  sparse representation features of the generalized orthogonal matching
  pursuit,'' \emph{Remote Sensing}, vol.~16, no.~17, 2024.

\bibitem{barbato2024ticino}
M.~P. Barbato, F.~Piccoli, and P.~Napoletano, ``Ticino: A multi-modal remote
  sensing dataset for semantic segmentation,'' \emph{Expert Systems with
  Applications}, vol. 249, p. 123600, 2024.

\bibitem{yan2024collection}
H.~Yan, R.~Zheng, S.~Mallela, R.~Russell, and O.~Kursun, ``Collection of a
  hyperspectral atmospheric cloud dataset and enhancing pixel classification
  through patch-origin embedding,'' \emph{Remote Sensing}, vol.~16, no.~17, p.
  3315, 2024.

\bibitem{raj2023investigating}
S.~P.~A. Raj, A.~Mishra, and H.~Sivaraman, ``Investigating the use of
  multi-sourced input data for time series algorithms applied to hyper spectral
  imagery,'' in \emph{International Conference on Data Science, Machine
  Learning and Applications}.\hskip 1em plus 0.5em minus 0.4em\relax Springer,
  2023, pp. 601--606.

\bibitem{A_3D_Deep_CNN_paper}
M.~Kanthi, T.~H. Sarma, and C.~S. Bindu, ``A 3d-deep cnn based feature
  extraction and hyperspectral image classification,'' in \emph{2020 IEEE India
  Geoscience and Remote Sensing Symposium (InGARSS)}, 2020.

\bibitem{RSVQA}
C.~Chappuis, V.~Zermatten, S.~Lobry, B.~Le~Saux, and D.~Tuia, ``Prompt-rsvqa:
  Prompting visual context to a language model for remote sensing visual
  question answering,'' in \emph{Proceedings of the IEEE/CVF Conference on
  Computer Vision and Pattern Recognition}, 2022, pp. 1372--1381.

\bibitem{HSI-BERT}
J.~He, L.~Zhao, H.~Yang, M.~Zhang, and W.~Li, ``Hsi-bert: Hyperspectral image
  classification using the bidirectional encoder representation from
  transformers,'' \emph{IEEE Transactions on Geoscience and Remote Sensing},
  vol.~58, no.~1, pp. 165--178, 2019.

\bibitem{segment}
I.~O. Sigirci and G.~Bilgin, ``Spectral-spatial classification of hyperspectral
  images using bert-based methods with hyperslic segment embeddings,''
  \emph{IEEE Access}, vol.~10, pp. 79\,152--79\,164, 2022.

\bibitem{knowledge}
L.~Mi, X.~Dai, J.~Castillo-Navarro, and D.~Tuia, ``Knowledge-aware text-image
  retrieval for remote sensing images,'' \emph{arXiv preprint
  arXiv:2405.03373}, 2024.

\bibitem{potato_paper}
J.~Lapajne, A.~Vojnović, A.~Vončina, and U.~Žibrat, ``Enhancing
  water-deficient potato plant identification: Assessing realistic performance
  of attention-based deep neural networks and hyperspectral imaging for
  agricultural applications,'' \emph{Plants}, vol.~13, no.~14, 2024.

\bibitem{blueberry_paper}
B.~Deng, Y.~Lu, and E.~Stafne, ``Fusing spectral and spatial features of
  hyperspectral reflectance imagery for differentiating between normal and
  defective blueberries,'' \emph{Smart Agricultural Technology}, vol.~8, 2024.

\bibitem{paddy_seed}
A.~A. Siam, M.~M. Salehin, M.~S. Alam, S.~Ahamed, M.~H. Islam, and A.~Rahman,
  ``Paddy seed viability prediction based on feature fusion of color and
  hyperspectral image with multivariate analysis,'' \emph{Heliyon}, vol.~10,
  no.~17, p. e36999, 2024.

\bibitem{wood_paper}
P.~P. Htun, M.~Boschetti, A.~Buriro, R.~Confalonieri, B.~Sun, A.~N. Htwe, and
  T.~Tillo, ``A lightweight approach for wood hyperspectral images
  classification,'' in \emph{2021 IEEE International Conference on Multimedia
  \& Expo Workshops (ICMEW)}, 2021, pp. 1--4.

\bibitem{BERT}
J.~Devlin, M.-W. Chang, K.~Lee, and K.~Toutanova, ``Bert: Pre-training of deep
  bidirectional transformers for language understanding,'' in \emph{Proceedings
  of NAACL-HLT}, 2019, pp. 4171--4186.

\bibitem{RoBERTa}
Y.~Liu, M.~Ott, N.~Goyal, J.~Du, M.~Joshi, D.~Chen, O.~Levy, M.~Lewis,
  L.~Zettlemoyer, and V.~Stoyanov, ``Ro{\{}bert{\}}a: A robustly optimized
  {\{}bert{\}} pretraining approach,'' 2020.

\bibitem{W2V}
T.~Mikolov, K.~Chen, G.~S. Corrado, and J.~Dean, ``Efficient estimation of word
  representations in vector space,'' in \emph{International Conference on
  Learning Representations}, 2013.

\end{thebibliography}

\end{document}